\documentclass[10pt,twocolumn,letterpaper]{article}

\usepackage{cvpr}              
\usepackage{booktabs}
\usepackage{multirow}
\usepackage{bm}
\usepackage{array}
\usepackage{algorithm}
\usepackage{algorithmic}
\usepackage{amsmath}
\usepackage{amsfonts}

\definecolor{cvprblue}{rgb}{0.21,0.49,0.74}
\usepackage[pagebackref,breaklinks,colorlinks,allcolors=cvprblue]{hyperref}

\def\paperID{5772} 
\def\confName{CVPR}
\def\confYear{2025}

\title{Expanding Event Modality Applications through a Robust CLIP-Based Encoder}

\author{
    Sungheon Jeong$^1$ \; Hanning Chen$^1$ \; Sanggeon Yun$^1$ \; Suhyeon Cho$^2$ \\ Wenjun Huang$^1$ \; Xiangjian Liu$^1$ \; Mohsen Imani$^1$\\[0.5em]
    $^1$University of California, Irvine \quad $^2$Pusan National University\\[0.5em]
}

\begin{document}
\maketitle
\begin{abstract}
    This paper introduces a powerful encoder that transfers CLIP’s capabilities to event-based data, enhancing its utility and expanding its applicability across diverse domains. While large-scale datasets have significantly advanced image-based models, the scarcity of comprehensive event datasets has limited performance potential in event modality. To address this challenge, we adapt CLIP’s architecture to align event embeddings with image embeddings, supporting zero-shot learning and preserving text alignment while mitigating catastrophic forgetting. Our encoder achieves strong performance in object recognition, with competitive results in zero-shot and few-shot learning tasks. Notably, it generalizes effectively to events extracted from video data without requiring additional training, highlighting its versatility. Additionally, we integrate this encoder within a cross-modality framework that facilitates interaction across five modalities—Image, Event, Text, Sound, and Depth—expanding the possibilities for cross-modal applications. Overall, this work underscores the transformative potential of a robust event encoder, broadening the scope and utility of event-based data across various fields. The code is available at: \url{https://github.com/EavnJeong/Event_Modality_Application}
\end{abstract}    
\vspace{-3.5mm}
\section{Introduction}
    \label{sec:intro}
        Event modality captures asynchronous changes in pixel brightness using event-based cameras ~\cite{gallego2020event, lichtsteiner2008128}, in contrast to traditional cameras that record full frames at regular intervals. This modality represents pixel-level changes in brightness, providing high temporal resolution, low latency, and reduced data redundancy. Due to these advantages, event modality has promising applications across various fields, particularly where rapid motion capture and real-time analysis are essential ~\cite{zheng2024eventdance, yao2024event, gallego2020event, zheng2023deep, chakravarthi2024recent}. However, event-based cameras capture only changes in pixel brightness, there exists a significant information gap compared to image. Consequently, models developed solely using event modality often face limitations in performance and scalability relative to image models ~\cite{gallego2020event, wu2023eventclip, zhou2024eventbind, chakravarthi2024recent, yang2023event, gu2021eventdrop}.

        The potential applications of event modality are currently constrained by the lack of a robust encoder, due to the limited availability of large datasets and the sparse information inherent in event data. Studies suggest that key visual elements are shared between event and image, offering a potential path to overcome these challenges~\cite{cho2023label, paredes2021back, rebecq2019events, yang2023event}. Nonetheless, these studies do not ensure effective performance on zero-shot tasks or extend to applications such as anomaly detection, which restricts their applicability across text and other specialized tasks in different modalities. Accordingly, we aim to leverage the processing power of CLIP~\cite{radford2021learning}, trained on large-scale datasets, to apply shared visual information to the event and develop a high-performance encoder. Generally, this knowledge transfer process occurs within the same modality~\cite{gupta2023cliptrans, cheng2024transfer, chefer2022image, pan2022st}; however, applying it across distinct modalities, such as from images to events, requires careful consideration~\cite{chen2023understanding, wang2023transferring, huang2023clip2point, patni2024ecodepth, zhou2023anomalyclip}. Without careful adaptation, there is a risk that CLIP could overfit to the new modality~\cite{zheng2023preventing, wang2023transferring, allgeuer2024unconstrained, wang2023improving}, forgetting its original capabilities of understanding. Therefore, it is essential to develop an encoder that captures unique features of event data while preserving the broad, image-based comprehension CLIP was initially trained to deliver. With these considerations in mind, our goal is to prevent model collapse and forgetting due to input gaps, maintaining CLIP’s foundational understanding while distinguishing between information that can and cannot be extracted from the event modality.
        
        If we can successfully transfer CLIP’s capabilities to event, we can leverage its text-alignment and zero-shot performance to broaden the applicability of event. This would allow the model to be used on new datasets or events generated from videos without additional training. Beyond just event-image tasks, this approach could open up possibilities for cross-modality tasks, such as multi-modal, classification, and generation involving other modalities like sound and depth. This expansion into cross-modality applications holds significant potential for advancing diverse fields requiring integrated data from multiple sensory sources.
        
        We structure CLIP in parallel to process both images and events independently, training each to be represented within a shared embedding space that enables alignment with textual representations. By mapping CLIP’s image embeddings to the same space as event embeddings, we transfer CLIP’s capabilities while extracting features specific to events. Additionally, we add a loss to prevent forgetting of image information, preserving zero-shot performance. Our approach allows for the transfer of CLIP's image-processing capabilities without model forgetting ~\cref{fig:5}, while simultaneously extracting relevant features from the event modality. Crucially, we exclude non-existent information in events, e.g. color, while successfully transferring learnable attributes like background and context to the event modality~\cref{fig:2}. Our model achieves state-of-the-art performance in object recognition and demonstrates meaningful results in few-shot scenarios. We further validate the encoder by applying it to video-extracted events, showcasing its practical applicability. Ultimately, we integrate the event encoder into a unified cross-modality model ~\cite{girdhar2023imagebind, zhu2023languagebind}, enabling interaction across five modalities (Image, Event, Text, Sound, Depth) and expanding the scope of event modality applications. Our contributions are as follows:

        \begin{itemize}
            \item
                We successfully transfer CLIP to the event modality, creating an event-image-text aligned model, achieving gains of +15.16\% in zero-shot, +18.91\% in 1-shot, and +7.35\% in fine-tuning over the state-of-the-art, using the same alignment method.
            \item
                Our model expands the applicability of event modality by using it for video anomaly detection, a task that has never been attempted before with event-based approaches.
            \item 
                We integrate the event encoder into a cross-modal framework~\cite{girdhar2023imagebind, zhu2023languagebind}, allowing for interaction across five modalities (Image, Event, Text, Sound, Depth) and significantly broadening the scope of applications for event modality.
        \end{itemize}
\section{Related Work}
    \label{Related Work}

    \noindent\textbf{Event Modality.}
        Event modality utilizes Dynamic Vision Sensors~\cite{lichtsteiner2008128} to encode changes in light intensity at the pixel level over time. Compared to traditional images, the information density of event data is significantly lower~\cite{wang2022exploiting, chakravarthi2024recent}, posing various challenges in handling event modality~\cite{gallego2020event, zheng2023deep}. Despite these differences, studies such as~\cite{cho2023label, paredes2021back, rebecq2019events} have shown that gray-scale images can be reconstructed from event data, indicating an overlap of critical information between the two modalities. However, the reconstructed information is primarily limited to pixel brightness~\cite{gallego2020event, wu2023eventclip}, lacking detailed information (e.g. color)~\cite{gallego2020event, chakravarthi2024recent}. There are also cases where large pre-trained models have demonstrated some ability to capture overlapping information~\cite{zhou2024eventbind, wu2023eventclip, wang2019ev, liu2022fast, kim2021n}. This highlights the potential for leveraging large pre-trained image model`s capabilities alongside the unique characteristics of event modality to extract meaningful features.

    \vspace{0.5mm}
    \noindent\textbf{Vision-Language Model.}
        Vision-language models, such as CLIP~\cite{radford2021learning}, have drawn significant attention for aligning images and text within a shared embedding space. This approach has been extended to various modalities, including point clouds~\cite{zhang2022pointclip, hess2024lidarclip, zeng2023clip2}, wave~\cite{wu2022wav2clip}, depth~\cite{auty2023learning, zhang2022can}, and audio~\cite{guzhov2022audioclip, shih2023speechclip}, with models like~\cite{girdhar2023imagebind, zhu2023languagebind} achieving simultaneous alignment across these modalities. While substantial progress~\cite{zhou2023clip, wu2023eventclip, zhou2024eventbind} has been made in integrating diverse modality, research in event-based modalities remains limited due to the absence of robust encoders and large datasets~\cite{wang2022exploiting, zhou2023clip}. Despite these challenges, efforts such as ~\cite{wu2023eventclip}, which uses feature adapters to aggregate temporal information and refine text embeddings, and ~\cite{zhou2024eventbind}, which introduces a module for tripartite alignment of image, event, and text embeddings, have emerged. These approaches highlight the need for continued development of powerful event encoders to incorporate recent cross-modality model.

    \vspace{0.5mm}
    \noindent\textbf{CLIP Modality Transfer.}
        While there are numerous applications leveraging CLIP~\cite{zhang2024vision, goel2022cyclip}, substantial research has focused on fine-tuning it using adapters~\cite{gao2024clip, zhang2022tip} and employing its representations in various learning tasks~\cite{zhong2022regionclip}. This approach is applied not only across different categories within the same modality~\cite{wang2023clipn, liu2023revisiting} but also between disparate modalities~\cite{chen2023understanding, kim2022transferring}. However, when the gap between two modalities is too large, there is a risk of CLIP losing its inherent capabilities~\cite{shi2023towards, zheng2023preventing}. Maintaining CLIP’s exceptional zero-shot performance while integrating it with other modalities requires careful handling~\cite{wang2023clipn, zheng2023preventing, gu2021open}. To address this, studies ~\cite{zheng2023preventing} have been developed to minimize catastrophic forgetting despite the differences between modalities. Building on this foundation, we aim to leverage CLIP to transition from image to event modality, preserving its capabilities while constructing a powerful encoder.
\newtheorem{proposition}{Proposition}

\section{Method}
    \label{Method}

    \begin{figure*}
        \centering
        \includegraphics[width=0.85\linewidth]{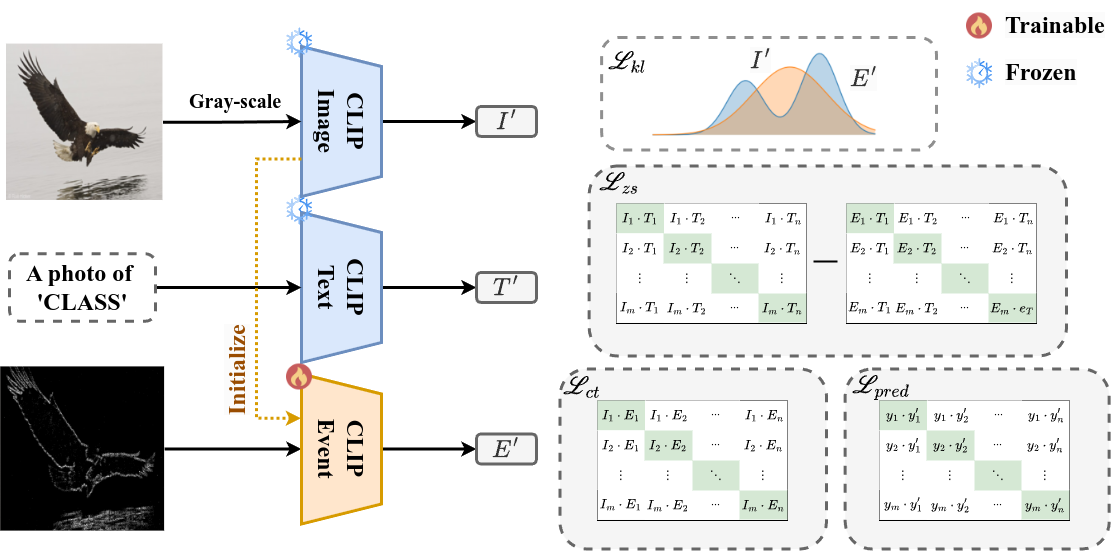}
    
        \caption{
            Overview of the proposed approach for aligning event and image representations within the CLIP framework. The image and event data are processed through separate encoders, with the image encoder \(f_I\) and text encoder \(f_T\) frozen, while the event encoder \(f_E\) is trainable. Various loss functions, including \(L_{\text{ct}}\), \(L_{\text{zs}}\), and \(L_{\text{kl}}\), ensure robust alignment across modalities and prevent collapse, facilitating the learning of shared features between events and images. \(L_{\text{pred}}\) is used only during the fine-tuning stage, where it provides direct supervision by aligning the prediction of \(f_E\), \(y^\prime\), with one-hot labels \(y\).
        }
        \label{fig:fig1}
    \end{figure*}
    
    In this work, we propose a novel approach to align event and image representations within the CLIP framework to enhance its capability in handling event-based data. To achieve this, we first investigate how to represent event modality effectively for CLIP, ensuring it retains as much temporal and spatial information as possible while aligning with CLIP's existing image encoder. Next, we address the challenge of aligning the event encoder with the image encoder to prevent the collapse of CLIP's pre-trained capabilities. We then detail our objective function, which leverages contrastive learning and includes additional mechanisms to maintain the integrity of CLIP’s zero-shot performance and stabilize learning through the incorporation of KL divergence and zero-shot consistency losses. Each component is thoroughly explained in the following sections.
    \subsection{Event Representation for CLIP}
        \label{Event Representation for CLIP}
        The event modality $E(x, y, t, p) = (E_x, E_y, E_t, E_p)$ captures frame-to-frame changes along the temporal axis ($t$), encoding the activated regions at spatial coordinates ($x$, $y$) in a digital format with polarity ($p$). Unlike previous studies such as~\cite{zhou2024eventbind, wu2023eventclip}, which aim to fully exploit the unique characteristics of $E$, our approach focuses on transferring CLIP's capabilities by encoding as much information as possible into a single-frame representation of $E$, making it comprehensible to CLIP. We aggregate event data across $t$ and $p$, expressed as $E(x, y) = \sum_{p}\sum_{t} E(x, y, t, p)$, and normalize it to construct $E$ as a one-channel gray-scale representation similar to $I$, as shown in ~\cref{eq:1}.

        \begin{equation}
            \label{eq:1}
            E = \frac{E(x, y)}{\max (E(x, y))+1}    
        \end{equation}

    \subsection{Event-Image Encoder Alignment}
        \label{Event-Image Encoder Alignment}
        Despite incorporating as much information as possible into $E$, there remains a significant disparity between $E$ and $I$. When training CLIP with $E$, this disparity can lead to a collapse, where the model loses its pre-existing capabilities due to the differences in the data. To address this, we utilize a trained CLIP model comprising the image encoder ($f_I$) and the text encoder ($f_T$), which serve as a reference to retain the original understanding. Additionally, we introduce an event encoder ($f_E$), initialized with $f_I$, to handle the $E$. The $f_I$ processes gray-scale images and, along with $f_T$, is frozen during training to consistently provide their pre-trained capabilities\cref{fig:fig1}.

    \subsection{Objective Function}
            In constructing the objective function, we consider the following key aspects. First, ensuring that the event embedding $E^\prime$ and the text embedding $T^\prime$ are well-aligned in the embedding space $\mathbb{R}^z$. Second, updating the $f_E$ while preserving the capabilities of the $f_I$ as much as possible. Lastly, aiming to build an encoder that effectively understands $E$.
            
        \vspace{0.5mm}
        \noindent\textbf{Contrastive Learning Approach.}
            While contrastive learning is a powerful tool to achieve above goals, directly training $E^\prime$ and $T^\prime$ through contrastive learning presents challenges. Specifically, $E^\prime$ may struggle to capture the encoding process of $f_I$, and focusing solely on matching $E^\prime$ with $T^\prime$ could lead to an $f_E$ that deviates from the interpretative process of $f_I$, tailoring itself exclusively to $E$. To address this, we employ contrastive learning between $E^\prime$ and $I^\prime$, allowing $f_E$ to directly observe $I^\prime$, and optimize using the InfoNCE~\cite{oord2018representation} as illustrated in \cref{eq:2}. Here, $E^\prime$ serves as the query set, while $I^\prime$ represents the key set. The encoder $f_E$ is optimized to make $E^\prime$ similar to the positive key $I^\prime_+$, while distancing it from all non-matching keys $I^\prime_-$. Since $I^\prime$ and $T^\prime$ are aligned in the latent space $\mathbb{R}^z$, this alignment facilitates the learning process, enabling $E^\prime$ and $T^\prime$ to become aligned as well.

            \begin{equation}
                L_{\text{ct}} = -\log \frac{\exp (E^\prime \cdot I^{\prime}_+ / \tau)}{\sum_{j=1}^N \exp(E^\prime \cdot I^\prime_j / \tau)}  
                \label{eq:2}
            \end{equation}

        \vspace{0.5mm}
        \noindent\textbf{Preventing Collapse with $\bm{L_{\text{zs}}}$.}
            However, this approach fails to prevent the collapse of CLIP caused by the significant differences between $E$ and $I$. This collapse manifests as an initial drop of accuracy immediately after training, followed by a gradual increase \cref{fig:5}. This indicates that instead of following the original process of $f_I$, $f_E$ focuses solely on mimicking the output of $f_I$, leading to the forgetting of CLIP’s zero-shot capabilities and other learned features. To address this, we propose incorporating $L_{\text{zs}}$ as suggested by~\cite{zheng2023preventing}. The $L_{\text{zs}}$ loss operates by using $I^\prime$ and $E^\prime$ to form a prediction logit matrix based on their similarities with $T^\prime$, treating $T^\prime$ as the label for both $E$ and $I$. This mechanism prevents overfitting to the unique characteristics of $E$, ensuring that CLIP retains its understanding of image features and extracts common characteristics across both modalities. Consequently, the event modality leverages CLIP’s comprehensive understanding of features, mitigating the risk of forgetting the $I$`s insights and preserving the capabilities that cannot be learned solely from $E$ due to its limited information. This approach allows the training to start from a reasonable accuracy level and progressively improve without significant loss \cref{fig:5}.

            \begin{align}
                L_{\text{zs}} = - \frac{1}{N} \sum_{i=1}^{N} \Bigg[ &\log \frac{\exp(E'_i \cdot T'_i / \tau)}{\sum_{j=1}^{M} \exp(E'_i \cdot T'_j / \tau)} \notag \\
                &+ \log \frac{\exp(I'_i \cdot T'_i / \tau)}{\sum_{j=1}^{M} \exp(I'_i \cdot T'_j / \tau)} \Bigg]
                \label{eq:3}
            \end{align}
    
            \begin{proposition}
                Let \( f_E \) be the query encoder, \( f_I \) the fixed key encoder, and \( f_T \) the text embedding module. The ZSCL (Zero-Shot Contrastive Learning) ~\cite{zheng2023preventing} objective aligns the query embeddings \( f_E(E) \) with the text embeddings \( f_T(T) \), and the key embeddings \( f_I(I) \) with \( f_T(T) \), through the above \cref{eq:3}: Only the parameters of \( f_E \) are updated during training, with the gradient computed as:
                
                \begin{align}
                    \nabla_{\theta_E} L_q = -\frac{1}{N} \sum_{i=1}^N \Bigg( 
                        & \frac{\nabla_{\theta_E} \exp(f_E(E_i) \cdot f_T(T_i) / \tau)}{\exp(f_E(E_i) \cdot f_T(T_i) / \tau)} \notag \\
                        &\hspace{-2em} - \frac{\sum_{j=1}^N \nabla_{\theta_E} \exp(f_E(E_i) \cdot f_T(T_j) / \tau)}{\sum_{j=1}^N \exp(f_E(E_i) \cdot f_T(T_j) / \tau)} 
                    \Bigg)
                \end{align}
                The parameter update for \( \theta_E \) follows a momentum-like behavior, aligning \( f_E \) with the fixed target space, where \( \theta_{\text{target}} \) represents the target parameter vector derived from the fixed key encoder or the text embedding module:
    
                \[
                    \theta_E \leftarrow m \cdot \theta_E + (1 - m) \cdot (\theta_{\text{target}} - \eta \cdot \nabla_{\theta_E} L_q)
                \]
                This update rule gradually aligns \( \theta_E \) with the target parameter \( \theta_{\text{target}} \) while adjusting in the direction of the current loss gradient at a rate determined by the learning rate \( \eta \), maintaining momentum \( m \).

            \end{proposition}

            In this proposition, we formalize the $L_{zs}$ objective, as introduced by~\cite{zheng2023preventing}, which aligns the query embeddings from the event encoder \( f_E \) with the text embeddings from \( f_T \), and the key embeddings from \( f_I \) with the same text embeddings. Uniquely, only the parameters of \( f_E \) are updated during training, maintaining consistency with the pre-trained key and text encoders. The gradient computation for \( L_q \) emphasizes maximizing the similarity between the query and text embeddings while reducing similarity with non-matching embeddings. The parameter update follows a momentum-like behavior, progressively aligning \( f_E \) with the fixed target space defined by \( f_I \) or \( f_T \). This strategy stabilizes the learning process, preventing abrupt changes in the learned embeddings and promoting robust alignment.

        \vspace{0.5mm}
        \noindent\textbf{KL Divergence Loss $L_{kl}$}
            \textbf{Losses.} The probability distribution alignment $L_{kl}$ proposed by ~\cite{yang2023event} demonstrated that aligning the embedding distributions of $E^\prime$ and $I^\prime$ is crucial for effectively understanding $E$, leading to build a strong model. We incorporate $L_{kl}(E^\prime||I^\prime) = \frac{1}{N}\sum_{i=1}^N E^\prime(i) \log \frac{E^\prime(i)}{I^\prime(i)}$ between $E^\prime$ and $I^\prime$ into the final objective function \cref{eq:4}.

        \vspace{0.5mm}
        \noindent\textbf{Final Objective Function:}
            The final objective function combines these losses:

            \begin{equation}
                L = L_{ct} + \alpha L_{zs} + L_{kl}
                \label{eq:4}
            \end{equation}

\section{Experiments}
    \label{Experiments}

    \begin{table*}
          \centering
          \begin{tabular}{@{}lcccccc@{}}
                \toprule
                Method & Reference & Zero-Shot & Labels & N-ImageNet & N-Caltech101 & N-MNIST \\
                \midrule
                \multicolumn{7}{@{}l@{}}{\textbf{Self-supervised pre-training methods}} \\
                HATS~\cite{sironi2018hats} & CVPR 2018 & x & x & 47.14 & 64.2 & 99.1 \\
                AsynNet~\cite{messikommer2020event} & ECCV 2020 & x & x & - & 74.5 & - \\
                EvS-S~\cite{li2021graph} & ICCV 2021 & x & x & - & 76.1 & - \\
                AEGNN~\cite{schaefer2022aegnn} & CVPR 2022 & x & x & - & 66.8 & - \\
                MEM~\cite{klenk2024masked} & WACV 2024 & x & x & 57.89 & 90.1 & - \\
                \midrule
                \multicolumn{7}{@{}l@{}}{\textbf{Transfer learning of image supervised pre-training methods}} \\
                RG-CNN~\cite{bi2019graph} & ICCV 2019 & x & o & - & 65.7 & 99 \\
                EST~\cite{gehrig2019end} & ICCV 2019 & x & o & 48.93 & 81.7 & - \\
                E2VID~\cite{rebecq2019events} & CVPR 2019 & x & o & - & 86.6 & 98.3 \\
                Matrix-LSTM~\cite{cannici2020differentiable} & ECCV 2020 & x & o & 32.21 & 84.31 & 98.9 \\
                DiST~\cite{kim2021n} & ICCV 2021 & x & o & 48.43 & 86.81 & - \\
                EventDrop~\cite{gu2021eventdrop} & IJCAI 2021 & x & o & - & 87.14 & - \\
                DVS-ViT~\cite{wang2022exploiting} & ICIP 2022 & x & o & - & 83 & - \\
                Event Camera Data Pre-training~\cite{yang2023event} & ICCV 2023 & x & o & 69.46 & 87.66 & - \\
                \midrule
                \multicolumn{7}{@{}l@{}}{\textbf{Transfer learning of image-text supervised pre-training methods}} \\
                EventCLIP~\cite{wu2023eventclip} (ViT-L-14) & - & o & o & 53.2 & 93.57 & - \\
                EventBind~\cite{zhou2024eventbind} (ViT-B-32) & ECCV 2024 & o & o & 42.94 & 93.74 & 99.26 \\
                EventBind~\cite{zhou2024eventbind} (ViT-L-14) & ECCV 2024 & o & o & 64.16 & 95.29 & 99.45 \\ \hline
                Ours (ViT-B-32) & - & o & o & 62.04 & 89.26 & 99.17 \\
                Ours (ViT-L-14) & - & o & o & \textbf{71.51} & \textbf{95.41} & \textbf{99.45} \\
                \bottomrule
          \end{tabular}
          \caption{
                Comparison of self-supervised and transfer learning methods across various datasets. The table reports Top-1 accuracy of object recognition on N-ImageNet, N-Caltech101, and N-MNIST. Our approach (ViT-B/32 and ViT-L/14) demonstrates superior performance, particularly on N-ImageNet and N-Caltech101, highlighting the effectiveness of leveraging pre-trained CLIP models for event-based object recognition.
            }
          \label{tab:1}
    \end{table*}

    \subsection{Experiment Settings}
        \noindent\textbf{Object Recognition.}
            We address the outcomes of zero-shot, few-shot, and finetuning approaches in object recognition. We pre-train our model using subsets of N-ImageNet mini~\cite{kim2021n} and ImageNet~\cite{deng2009imagenet}, consisting of 80\% of the classes and refer to this as N-ImageNet in the following. The remaining 20\% of N-ImageNet classes, along with the entire N-Caltech~\cite{orchard2015converting} and N-MNIST~\cite{orchard2015converting} datasets, are used for evaluation. As described in \cref{eq:1}, the event $E$ undergoes preprocessing. For N-Caltech and N-MNIST, clamping is applied along the time axis prior to normalization. This process limits the gray-scale representation and prevents over-representation caused by extended time steps, thereby mitigating out-of-distribution (OOD) issues.
            
            We utilize two pre-trained versions of CLIP, ViT-B/32 and ViT-L/14. The prompt (\(T\)) for \( f_T \) is set as "$a\; photo\; of\; \{class\}$", and \( I \) is processed by \( f_I \) after applying gray-scaling. For training, we prepare pairs using the trained CLIP model along with the N-ImageNet and ImageNet datasets. All encoders, except for \( f_E \), are kept frozen. The model is trained for 200 epochs with a learning rate of 1e-6. The temperature parameters \( \tau \) in \cref{eq:3} and \(\tau\) in \cref{eq:2} are initialized to 2 and 1, \( \alpha \) in \cref{eq:4} is set to 0.1.

            Using the pre-trained model, we fine-tune it by using $L_{\text{pred}}$ \cref{eq:5} is used during fine-tuning to align the predictions $y^\prime$ from $f_E$ with the one-hot labels $y$, using cross-entropy loss for supervision. For few-shot learning, we randomly sample $N$ data points per class for training.
            \begin{equation}
                L_{\text{pred}} = -\frac{1}{N} \sum_{i=1}^{N} \sum_{c=1}^{C} y_i^{(c)} \log y^\prime_{i^{(c)}} 
                \label{eq:5}
            \end{equation}
            
        \vspace{0.5mm}
        \noindent\textbf{BaseLine.}
            We compare our results with EventCLIP~\cite{wu2023eventclip} and EventBind~\cite{zhou2024eventbind} in both zero-shot and few-shot evaluations. For a fair comparison, we restrict the evaluation on N-ImageNet to classes that were not used during pre-training. The final fine-tuning results are compared against various baselines to assess the overall accuracy.
            
\begin{figure}
    \centering
    \includegraphics[width=0.9\linewidth]{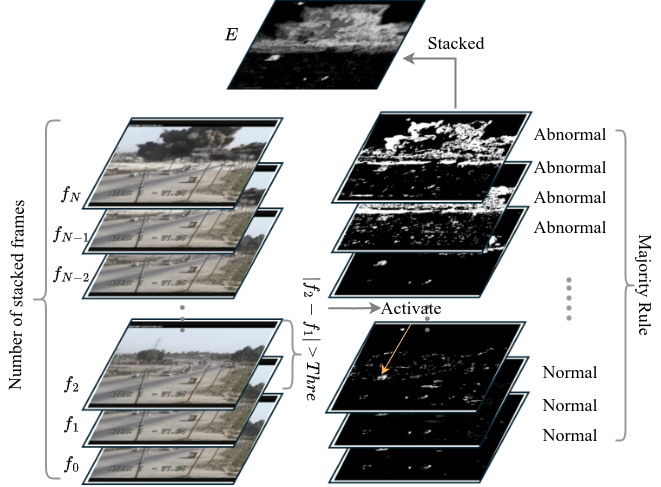}

    \caption{
        Extracting events from video frames. The differences between frames are activated based on threshold, and the resulting events are sequentially stacked to generate \( E \) from (\(f_0 \sim f_n\)). where \( N \) denotes the total number of frames in the stack.
    }
    \label{fig:3}
\end{figure}

        \vspace{0.5mm}
        \noindent\textbf{Event-based Video Anomaly Detection.}
            For event-based Video Anomaly Detection (VAD), we use the UCFCrime~\cite{sultani2018real}, XD-Violence~\cite{sultani2018real}, and Shanghaitech~\cite{luo2017revisit} datasets to extract events. The process begins by determining event pixels based on the frame-to-frame difference, using a pixel value threshold of 25 of 255. Specifically, if the difference between corresponding pixel values in consecutive frames exceeds this threshold, it is considered an event activated pixel. Each event is built using a sequence of 16 consecutive frames, collectively representing a single event instance \cref{fig:3}. Following the construction of events, we assign labels to each event by performing majority voting~\cite{zhou2018brief} across the frames in the event sequence, assigning a label of 0 for normal and 1 for abnormal behavior as similar to weakly supervised video anomaly detection~\cite{tian2021weakly}. 
            
        \vspace{0.5mm}
        \noindent\textbf{Event Retrieval.}
            For event retrieval, we integrate our encoder with ImageBind~\cite{girdhar2023imagebind} to enable zero-shot retrieval for event-sound and event-depth. To align the embedding spaces of our encoder and ImageBind’s, we incorporate an adapter layer, designed as a single-layer module. This adaptation extends our model's modalities (event, image, text) to include ImageBind's sound and depth modalities. In the event, image, and text modalities, we utilize the N-Caltech dataset~\cite{orchard2015converting}, while the ESC-50 dataset~\cite{piczak2015esc} is used for sound, and DENSE~\cite{hidalgo2020learning} is employed for depth.
        
\begin{table*}
    \centering
    \begin{tabular}{cccc|ccc|ccc}
        \toprule
        & \multicolumn{3}{c}{N-ImageNet} & \multicolumn{3}{c}{N-Caltech101} & \multicolumn{3}{c}{N-MNIST} \\ \cline{2-10}
        & EventCLIP & EventBind & Ours & EventCLIP & EventBind & Ours & EventCLIP & EventBind & Ours \\
        \hline
        0-shot & 22.74 & 11.52 & \textbf{37.9} & 58.82 & 61.8 & \textbf{66.34} & 48.72 & \textbf{56.81} & 46.95 \\
        1-shot & 25.39 & 17.6 & \textbf{44.3} & 75.82 & 74.96 & \textbf{79.43} & 74.62 & \textbf{74.64} & 69.43 \\
        2-shot & 29.44 & 21.7 & \textbf{45.91} & 78.86 & 79.78 & \textbf{81.91} & 82.78 & 82.89 & \textbf{83.33} \\
        5-shot & 35.43 & 28.1 & \textbf{50.4} & 83.57 & 84.49 & \textbf{85.01} & 93.78 & 94.44 & \textbf{94.23} \\
        \bottomrule
    \end{tabular}
    \caption{
        Comparison of zero-shot \& few-shot accuracy across N-ImageNet, N-Caltech101, and N-MNIST datasets for object recognition.
    }
    \label{tab:2}
\end{table*}

\begin{table}
    \centering
    \begin{tabular}{@{}lccc@{}}
        \toprule
        Method      & UCFCrime & XD-Violence & Shanghai    \\ \midrule
        \multicolumn{4}{@{}l@{}}{\textbf{Zero-shot text based learning methods}} \\
        CLIP        & 58.63    & 27.21       & 49.17       \\
        Ours & 64.14     & 57.65           & 54.41           \\ \bottomrule
    \end{tabular}
    \caption{
        VAD performances of zero-shot text based learning methods. All datasets use AUC as evaluation metric.
    }
    \label{tab:3}
\end{table}

    \subsection{Object Recognition}
        \vspace{0.5mm}
        \noindent\textbf{Zero \& Few-Shot.}
            Our pre-trained model effectively extracts \( T \) from \( E \), retaining CLIP's zero-shot capabilities while establishing a robust encoder aligned with self-supervised learning objectives. This approach yields superior zero-shot and few-shot performance compared to existing models, as shown in \cref{tab:2}. Specifically, in image-text alignment tasks, our method achieves performance improvements of \( +17.16\% \) and \( +4.54\% \) over the baseline, with further gains of \( +18.89\% \) and \( +5.47\% \) in the 1-shot setting. These results confirm the successful preservation and transfer of the original CLIP model's zero-shot functionality within our framework.
        
        \vspace{0.5mm}
        \noindent\textbf{Fine-tuning.}
            Using our pre-trained model, we conducted fine-tuning to perform supervised learning, achieving state-of-the-art performance across N-ImageNet, N-Caltech, and N-MNIST. Notably, in N-ImageNet, our image-text alignment method outperformed baselines with the same backbone sizes, achieving improvements of \( +19.08\% \) with ViT-B/32 and \( +7.35\% \) with ViT-L/14. Additionally, our approach outperformed conventional supervised learning methods that do not employ image-text alignment by approximately \( +2.05\% \) in N-ImageNet \cref{tab:1}.
    
    \begin{figure}
        \centering
        \includegraphics[width=0.9\linewidth]{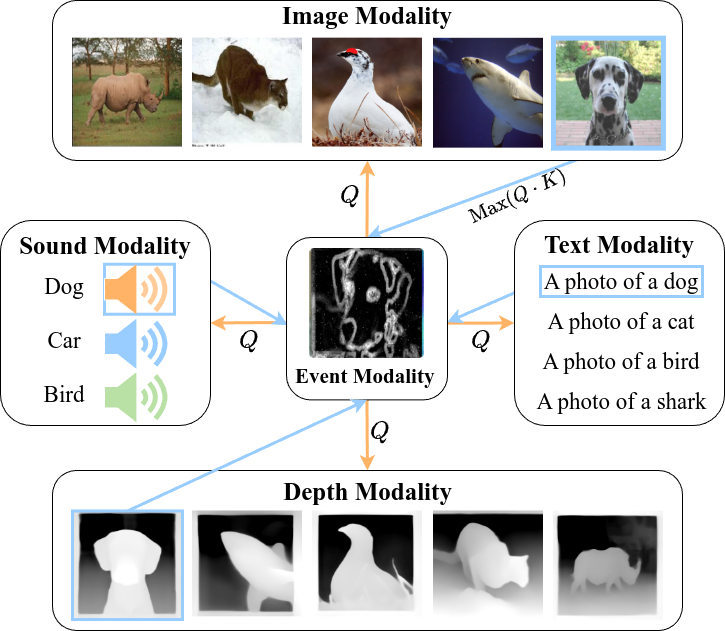}
    
        \caption{
            Event retrieval process across different modalities (Image, Text, Sound, Depth) using the event as query. The query event calculates the maximum similarity with each modality’s key embedding, returning the key modality with the highest similarity score.
        }
        \label{fig:4}
    \end{figure}
    
    \subsection{Event-based Video Anomaly Detection.}
        In this section, we demonstrate the utility of the event modality by extracting events from a video dataset (rather than an event-specific dataset) to perform VAD \cref{fig:3}. Our approach consistently outperforms the traditional image-text alignment-based zero-shot prediction model, CLIP, across all three datasets. Notably, in the XD-Violence dataset, we observe a performance improvement of +27.44, as shown in \cref{tab:3}. This finding is significant as it suggests that, even with limited but critical information (e.g. events) rather than the extensive detail available in conventional images, we can enhance performance on tasks like anomaly detection. Furthermore, our findings suggest that integrating the model with additional training, weak-supervised learning frameworks, or enhanced event-processing modules could potentially improve performance. This underscores the potential of the event modality to expand into previously unexplored areas.
        
    \subsection{Event Retrieval}
        In this section, we evaluate the ability of our model’s event embeddings to interact with other modalities by conducting zero-shot retrieval tasks from event to image, text, sound, and depth \cref{fig:4}. We measure by Recall@1, Recall@5, mAP@1, mAP@5, and Mean Reciprocal Rank (MRR). Without additional training, our model achieves Recall@1 scores of 71.03 and 61.15 for Event-Image and Event-Text retrieval, respectively, and Recall@10 scores of 96.09 and 93.10. Similar performance is observed for Image-Event and Text-Event retrieval. Furthermore, in the event-to-sound retrieval task, the model achieves an MRR of 82.90, and in the event-to-depth retrieval, it achieves an MRR of 62.99 \cref{tab:4}. These results indicate that our model can be integrated into existing cross-modal frameworks without training, demonstrating significant robustness and potential for broader applicability in multi-modal contexts.

    \begin{table}[t]
        \centering
        \begin{tabular}{lcccc}
            \toprule
            \textbf{Event to} & \textbf{Image} & \textbf{Text} & \textbf{Sound} & \textbf{Depth} \\
            \midrule
            Recall@1 & 71.03 & 61.15 & 55.37 & 52.20 \\
            Recall@5 & 90.46 & 87.70 & 82.77 & 78.46 \\
            Recall@10 & 96.09 & 93.10 & 90.11 & 88.46 \\
            mAP@1 & 71.03 & 61.15 & 55.37 & 51.12 \\
            mAP@5 & 79.03 & 71.88 & 66.07 & 63.01 \\
            MRR & 79.78 & 72.59 & 67.04 & 65.30 \\
            \midrule
            \textbf{Event from} & \textbf{Image} & \textbf{Text} & \textbf{Sound} & \textbf{Depth} \\
            \midrule
            Recall@1 & 72.76 & 56.84 & 74.01 & 51.38 \\
            Recall@5 & 91.26 & 84.48 & 95.76 & 75.54 \\
            Recall@10 & 95.57 & 92.70 & 100.00 & 83.07 \\
            mAP@1 & 72.76 & 56.84 & 74.01 & 50.07 \\
            mAP@5 & 80.10 & 67.86 & 82.35 & 61.74 \\
            MRR & 80.69 & 68.96 & 82.90 & 62.99 \\
            \bottomrule
        \end{tabular}
        \label{tab:4}
        \caption{Recall, mAP, and MRR for Event to and Event from other modalities}
    \end{table}
        
    \subsection{Ablations}
\begin{figure}
    \centering
    \includegraphics[width=1.0\linewidth]{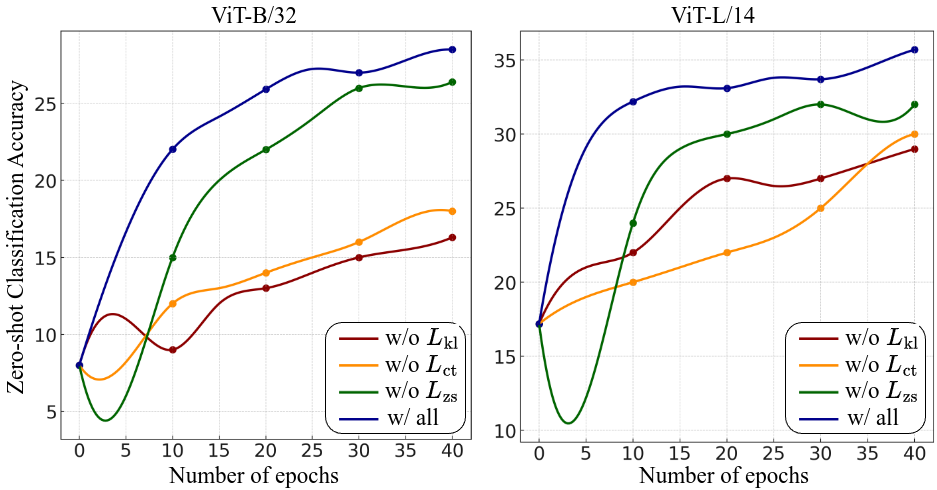}

    \caption{
        Zero-shot accuracy on unseen classes during N-ImageNet pre-training, measured for different loss configurations. The figure illustrates how the inclusion or exclusion of each loss component affects performance on unseen classes.
    }
    \label{fig:5}
\end{figure}

        We perform an ablation study on the composition of our objective function by removing each of the three components \( L_{\text{ct}} \), \( L_{\text{zs}} \), and \( L_{\text{kl}} \) from \cref{eq:4} in turn, and observing the zero-shot classification accuracy during pre-training. Notably, omitting \( L_{\text{zs}} \) leads to a decrease in accuracy, which may be linked to a form of forgetting, ultimately resulting in lower performance compared to using all loss components. The absence of \( L_{\text{ct}} \) and \( L_{\text{kl}} \) also impacts performance, particularly for the smaller ViT-B architecture compared to ViT-L. This finding suggests that the full loss configuration has a greater impact on smaller architectures, demonstrating its effectiveness for optimal CLIP transfer and improved performance \cref{fig:5}.

\begin{figure*}
    \centering
    \includegraphics[width=0.80\linewidth]{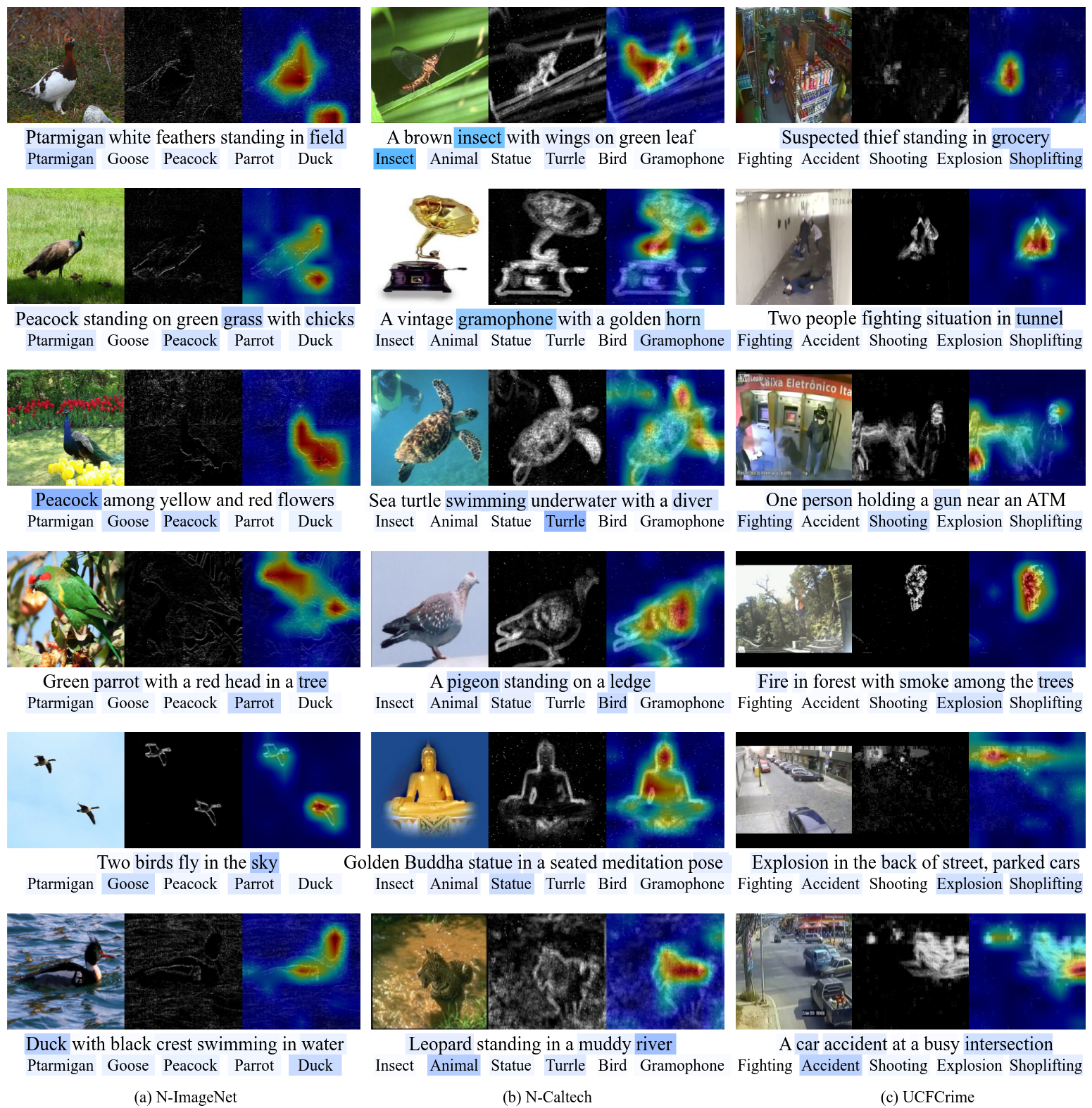}

    \caption{
        Visualize grad-based attention maps and text relevance scores with our pre-trained model (a) N-ImageNet unseen classes, (b) N-Caltech, (c) UCFCrime following the method in ~\cite{chefer2021generic}.
    }
    \label{fig:2}
\end{figure*}

    \subsection{Discussions}
        \noindent\textbf{Event data preprocessing.}
            Our study aims to transfer the capabilities of CLIP to build a robust encoder for \( E \). To this end, we designed inputs that are compatible with CLIP’s architecture, as shown in \cref{eq:1}. However, this approach may overlook certain features of \( t \) and \( p \) that are crucial in other contexts. Incorporating modules and preprocessing methods, such as those proposed in \cite{wu2023eventclip, zhou2024eventbind}, could potentially enhance performance. Additionally, by using a fixed text prompt, we may have missed opportunities to explore text prompts specifically suited for events; integrating prompt learning methods could further improve our results. 

        \noindent\textbf{Visualize the understand of the model.}
            We demonstrate the successful transfer of CLIP's capabilities through attention maps and text relevance scores. \cref{fig:2} displays the attention maps and text relevance produced by our model. The first row illustrates the model's understanding of sentence elements by showing text relevance to the description of each RGB image, while the second row represents the correlation with text (class names). Higher relevance is indicated by shades closer to cyan, while lower relevance is indicated by shades to white. Our model successfully captures high relevance even for untrained background elements and effectively excludes color-related associations.

        \noindent\textbf{Zero-shot performance.}
            We performed zero-shot evaluations across object recognition, anomaly detection, and event retrieval, achieving strong results in most cases. However, the N-MNIST dataset showed relatively low zero-shot performance, likely due to the representation of MNIST images differing substantially from typical images. Fine-tuning aligns the model with these distinctive MNIST features, enabling it to achieve state-of-the-art performance. Thus, we anticipate that fine-tuning our pre-trained model will similarly enhance performance across other tasks.

        \noindent\textbf{Limitation.}
            While our model achieves state-of-the-art performance in object recognition, its performance still lags behind image model. Additionally, although our experiments demonstrate the potential for applying the model to anomaly detection and event retrieval, further research is needed to fully harness this potential and achieve optimal performance. We hope that these findings will inspire future research to refine and expand upon our approach for even greater efficacy across diverse applications.
\section{Conclusion}
    \vspace{-0.1mm}
    \label{Conclusion}
    We leverage the shared and distinct features of CLIP's image and event modalities to transfer knowledge from images to events, focusing on information extractable from events. By preventing potential forgetting during this transfer, we extend zero-shot and text alignment capabilities acquired from large datasets to the event modality, achieving significant performance gains in object recognition. This demonstrates that events generated from video can also be effectively processed, showcasing the integration of event modality into a unified modality framework and ultimately expanding the applicable domain of event modality.
    
    \noindent\textbf{Broader impact.}
        Our model establishes strong zero-shot capabilities, aligning event-image-text modalities and expanding the scope of event to interact with other modalities. We expect that the event can be utilized in various tasks, either as standalone representations of video or event data, or in combination with other modalities.

\section*{Acknowledgment}
    This work was supported in part by the DARPA Young Faculty Award, the National Science Foundation (NSF) under Grants \#2127780, \#2319198, \#2321840, \#2312517, and \#2235472, the Semiconductor Research Corporation (SRC), the Office of Naval Research through the Young Investigator Program Award, and Grants \#N00014-21-1-2225 and \#N00014-22-1-2067, Army Research Office Grant \#W911NF2410360. Additionally, support was provided by the Air Force Office of Scientific Research under Award \#FA9550-22-1-0253, along with generous gifts from Xilinx and Cisco.

{
    \small
    \bibliographystyle{ieeenat_fullname}
    \bibliography{main}
}

\end{document}